\documentclass[10pt, a4paper]{article}
\usepackage{lrec2022} 
\usepackage{multibib}
\usepackage [utf8]{inputenc}
\newcites{languageresource}{Language Resources}
\usepackage{graphicx}
\usepackage{tabularx}
\usepackage{soul}

\usepackage{listings}
\usepackage{titlesec}
\titleformat{\section}{\normalfont\large\bfseries\center}{\thesection.}{1em}{}
\titleformat{\subsection}{\normalfont\SmallTitleFont\bfseries\raggedright}{\thesubsection.}{1em}{}
\titleformat{\subsubsection}{\normalfont\normalsize\bfseries\raggedright}{\thesubsubsection.}{1em}{}
\renewcommand\thesection{\arabic{section}}
\renewcommand\thesubsection{\thesection.\arabic{subsection}}
\renewcommand\thesubsubsection{\thesubsection.\arabic{subsubsection}}

\usepackage{booktabs, multirow}

\usepackage{epstopdf}
\usepackage[utf8]{inputenc}

\usepackage{hyperref}
\usepackage{xstring}

\usepackage{color}

\title{MuLD: The Multitask Long Document Benchmark}

\name{G Thomas Hudson, Noura Al Moubayed}

\address{Durham University \\
 Department of Computer Science, Durham University, Durham, United Kingdom \\
         \{g.t.hudson, noura.al-moubayed\}@durham.ac.uk\\}

\abstract{
The impressive progress in NLP techniques has been driven by the development of multi-task benchmarks such as GLUE and SuperGLUE. While these benchmarks focus on tasks for one or two input sentences, there has been exciting work in designing efficient techniques for processing much longer inputs. In this paper, we present MuLD: a new long document benchmark consisting of only documents over 10,000 tokens. By modifying existing NLP tasks, we create a diverse benchmark which requires models to successfully model long-term dependencies in the text. We evaluate how existing models perform, and find that our benchmark is much more challenging than their `short document' equivalents. Furthermore, by evaluating both regular and efficient transformers, we show that models with increased context length are better able to solve the tasks presented, suggesting that future improvements in these models are vital for solving similar long document problems. We release the data and code for baselines to encourage further research on efficient NLP models.
 \\ \newline  }

\begin{document}

\maketitleabstract

\section{Introduction}

Pretrained language models have been highly influential in Natural Language Processing (NLP), leading to state-of-the-art results across a wide range of tasks. Based on the transformer architecture, these language models have shown capable at text classification, question answering, and translation among many other NLP problems.

The rise of pretrained language models in NLP has been driven by influential benchmarks such GLUE \cite{Wang_2018}, and SuperGLUE \cite{Wang_2019}, which combine multiple existing datasets to provide a standardised evaluation of general-purpose Natural Language Understanding (NLU) approaches. Similar benchmarks have been created to evaluate multilingual approaches \cite{Yao_2021}, and other types of NLP task such as Natural Language Generation (NLG) \cite{Liu_2021}.

However, a key component of the success of the transformer model - self-attention - is also a major limitation when it comes to processing longer sequences. Comparing each token to all other tokens in the previous layer yields a $O(n^2)$ complexity, limiting the ability of standard transformers only a few hundred or thousand tokens on standard hardware. With many real world tasks involving the need to process documents in the range of tens of thousands of tokens, this is an important problem to solve.

\begin{figure}[t]
    \centering
    \includegraphics[width=1.0\linewidth]{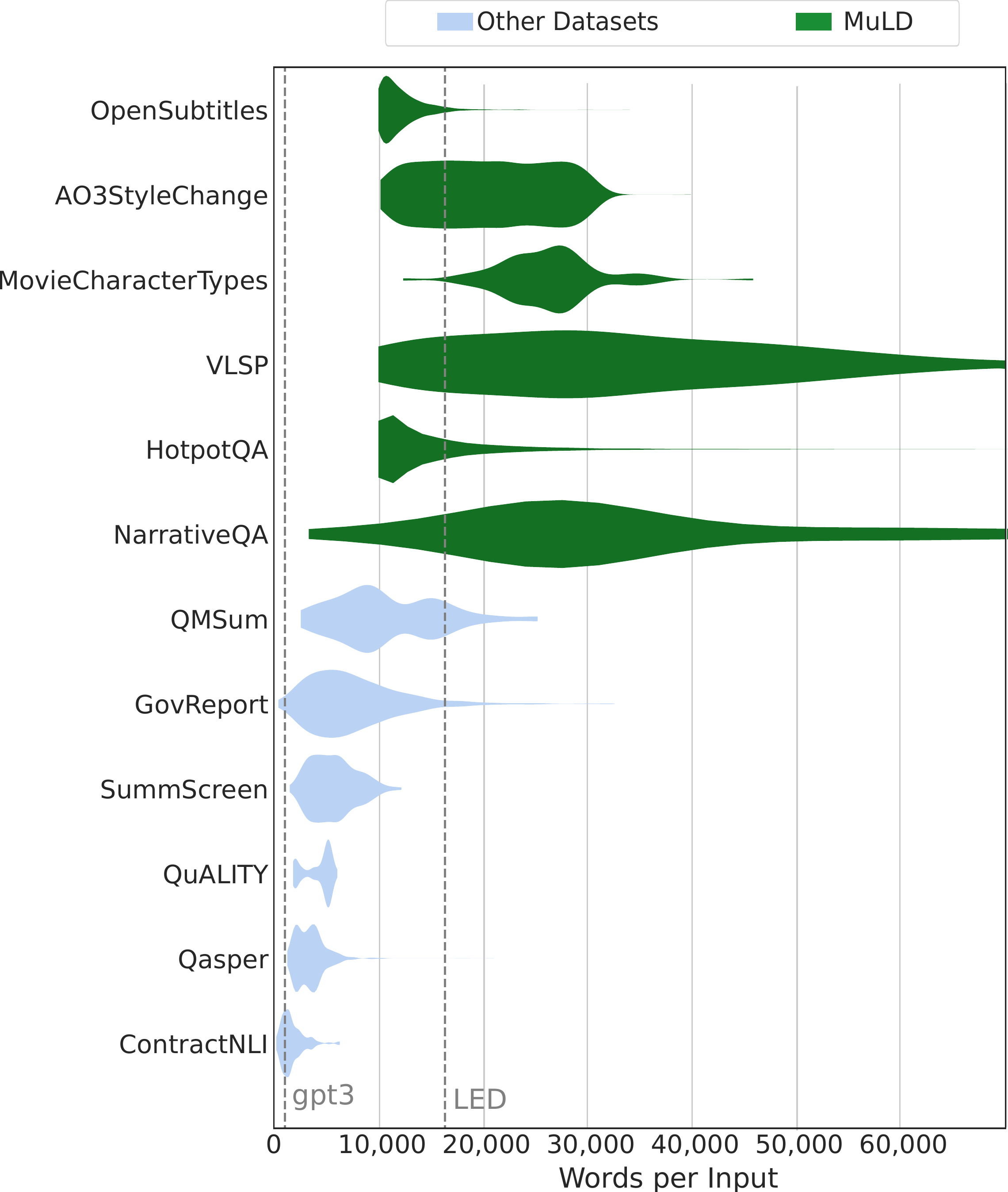}
    \caption{\centering Comparison of lengths of multitask long document benchmarks. The maximum input length of gpt3 and Longformer (LED) are included for comparison.}
    \label{fig:benchmark_comparison}
\end{figure}
Recently, approaches such as Longformer \cite{Beltagy_2020}, Reformer \cite{Kitaev_2020}, and Linformer \cite{Wang_2020} have explored techniques for improving the efficiency of transformers, allowing them to operate on much longer sequences. However, these have mainly been evaluated on artificial datasets, with limited evaluation on real-world data.

There has been some work on creating long-text datasets. A notable example is NarrativeQA \cite{Kocisky_2018} which challenges models to answer questions about the plot of entire novels and movie scripts which have an average length of around 60,000 tokens - well beyond the input size of current typical efficient transformer models. There has however been little work in developing benchmarks of this length across a wide variety of NLP tasks.

Many `long document' benchmarks use datasets consisting of at most a few thousand tokens (\autoref{fig:benchmark_comparison}). Notably, the Long-Range Arena \cite{Tay_2021LongRangeArena} uses a maximum length of 8K tokens, and the QuALITY becnhmark \cite{Pang_2021} uses a maximum of 6K tokens. In this paper, we argue that these lengths are more akin to short essays, and the common usage of the term `long document' would imply documents in the tens of thousands of tokens - similar in length to novels.

It is to this end that we present MuLD: The \textbf{Mu}ltitask \textbf{L}ong \textbf{D}ocument benchmark. This is a set of six long document tasks where each input is at least 10,000 tokens, spanning a range of dataset sizes, genres, and formulations designed specifically to test the ability of different approaches to model long-term dependencies in real world text. The datasets are formed by filtering, extending, or modifying existing NLP datasets.

We create baseline results for both standard approaches (T5) and efficient transformer methods (Longformer), finding that Longformer's extended input context allows it to perform better. 

The MuLD dataset is available at \url{www.github.com/ghomasHudson/muld}.

\section{Related Work}

\paragraph{Benchmarks}
The influence for many NLP benchmarks is the success of GLUE which challenged models to solve 9 language understanding tasks including question answering, coreference resolution, and sentiment analysis. The success of GLUE resulted in advancement to the point where a successor with more challenging tasks was required: SuperGLUE. This approach has been replicated both across other languages \cite{Yao_2021}, as well as other task domains such as NLG \cite{Liu_2021}.

Recently there have been some attempts to design benchmarks which test the ability of models to understand long documents.

The Long Range Arena was designed as a benchmark for evaluating the performance of efficient transformers on long input sequences \cite{Tay_2021LongRangeArena}. This benchmark challenges models to perform 6 synthetic and real-world multitmodal tasks on documents with up to 16,000 tokens. Howerver, these tasks are all forms of classification, and as noted by \newcite{Shaham_2022scrolls} one of the only two NLP tasks: LRA, uses byte tokenization as a way of artificially increasing the token count while having a much smaller number of words.

\newcite{Guan_2021} introduced a benchmark of long document Chinese tasks based on short stories. Four tasks were set including cloze tests, sentence position prediction, plot completion, and story generation. In contrast to this, we focus on English text, more conventional 'real-world' tasks, and crucially on documents with a minimum length of 10,000 tokens which we consider to be true 'long documents'.

The QuALITY benchmark \cite{Pang_2021}, was designed as a multiple-choice question answering dataset which selects questions which can't be answered by briefly skimming the text. Again the maximum document length used: 6,000, is much shorter than the minimum length used in our work of 10,000 tokens.

The most notable recent development is SCROLLS: Standardized CompaRison Over Long Language Sequences \cite{Shaham_2022scrolls} which used 7 long document tasks to create a benchmark with a range of input lengths with the median for most tasks (excluding the very long NarrativeQA) falling between 1000 and 10,000 words - which we argue is still too short to be a reliable evaluation of the longer transformer models such as the LED Longformer variant. Additionally, unlike our benchmark which has 5 different types of task, SCROLLS focuses on question answering, summarization and NLI. This limits the type and length of the expected outputs to only short sentences or paragraphs. In contrast, MuLD explores outputs lengths from single words all the way up to outputs that are an equivalent length to the input.

\paragraph{Long Document Models}
There have been numerous attempts to improve both the memory footprint and computational cost of transformers, thus allowing the use of longer inputs. One way of tackling the high complexity of full attention is to make the attention sparse. This can be done by chunking the input sequence into fixed blocks \cite{qiu-etal-2020-blockwise}, applying strided windows to the attention matrix, or some method of learning which tokens to attend to \cite{Kitaev_2020}. Other models such as transformer-XL \cite{Dai_2019transformerxl} and Infty-former \cite{Martins_2021infty} solve this problem by augmenting the model with a memory mechanism \cite{Katharopoulos_2020transformersarernns}. A form of the kernel trick can be used as well to reduce the complexity. All these techniques are effective in optimising the memory and computation usage of transformer models, but there is little analysis of how these techniques effect the ability of models to solve a wide variety of NLP tasks, which we seek to solve. 

For a detailed overview of efficient transformers, see \newcite{Tay_2020EfficientTA}.

\section{The MuLD Benchmark}

\begin{table*}[]
    \centering
    \begin{tabular}{lllcc}
    \toprule
    Dataset     & task & metrics &  \# documents & avg. \# tokens 
    \\
    \midrule

    \textbf{NarrativeQA} & Question Answering & bleu, rouge, meteor & \\ 
    Train & & & 32,159 & 91,938\\
    Valid & &  & 3,461 & 90,832\\
    Test & &  & 10,261 & 88,579\\
    \textbf{HotpotQA} & Question Answering & bleu, rouge, meteor & \\
    Train & & & 87,752 & 23,775\\
    Valid & & & 7,405 & 22,669\\
    \textbf{AO3 Style Change detection} & Style change detection & F1-score & \\
    Train & & & 6,354 & 29,657\\
    Valid & & & 705 & 29,304\\
    Test & &  & 2,352 & 30,502 \\
    \textbf{Movie Character Types} & Classification & F1-score\\
    Train & & & 167 & 44,640\\
    Test & & & 86 & 48,165\\
    \textbf{Very Long Scientific Papers} & Summarization & bleu, rouge, meteor\\
    Test & & & 482 & 57,473\\
    \textbf{Open Subtitles} & Translation & bleu, rouge, meteor\\
    Train & & & 4,252 & 12,330\\
    Test & & & 1,385 & 18,932\\
    \bottomrule
    \end{tabular}
    \caption{MuLD data statistics}
    \label{tab:dataset_summary}
\end{table*}
The MuLD benchmark is based on six long document NLP tasks, which span a wide range of domains. 

\subsection{Desiderata}
We pick these tasks based on the following principles: \paragraph{(1) Long input size:} While many benchmarks use `long-text' to mean inputs of a few thousand tokens, we consider `true' long documents to be in the tens of thousands of tokens long. This more closely matches the common usage of this term, where an essay may be a few thousand words long and considered fairly `short', while common everyday examples of long documents such as novels and reports may be 50,000-100,000 words in length. It is for this reason that we only include documents over 10,000 tokens in our benchmark, with many documents exceeding 100,000 tokens. \paragraph{(2) Variety of dependency on the input:} Within long document tasks, some may require understanding more of the input than others - either analysing the whole text or just using relevant sections. For example in summarization, a model must have a holistic overview of the document, while in other tasks such as question answering, the answer can be often be given by referencing just a few sections of the document. In reality, most tasks fall sommewhere between these two extremes and we endeavour to capture a variety of input dependency in our benchmark. \paragraph{(3) Variety of output length:} While the input length should be long, there is a range of different possible output lengths ranging from a single word classification label, a short answer, all the way up to an output of equivalent length to the input. \paragraph{(4) Existing Task Formulation} We don't seek to invent new types of task, but instead use proven tasks which are already agreed as being challenging for regular `short' transformer models. Instead, the tasks used are created by either filtering existing datasets, expanding the length of existing datasets with additional text, or replicating the methodology of existing datasets with longer source text. \paragraph{(5) Easily Evaluated} We pick tasks where the performance can be easily measured with multiple automatic metrics. We also acknowledge that for some tasks, the current metrics used on short documents don't fully capture the challenges that long document evaluation poses.

\subsection{The Tasks}

The six long document tasks are described below and the summarized in \autoref{tab:dataset_summary}.

\begin{figure*}
    \centering
    \includegraphics[width=1.0\textwidth]{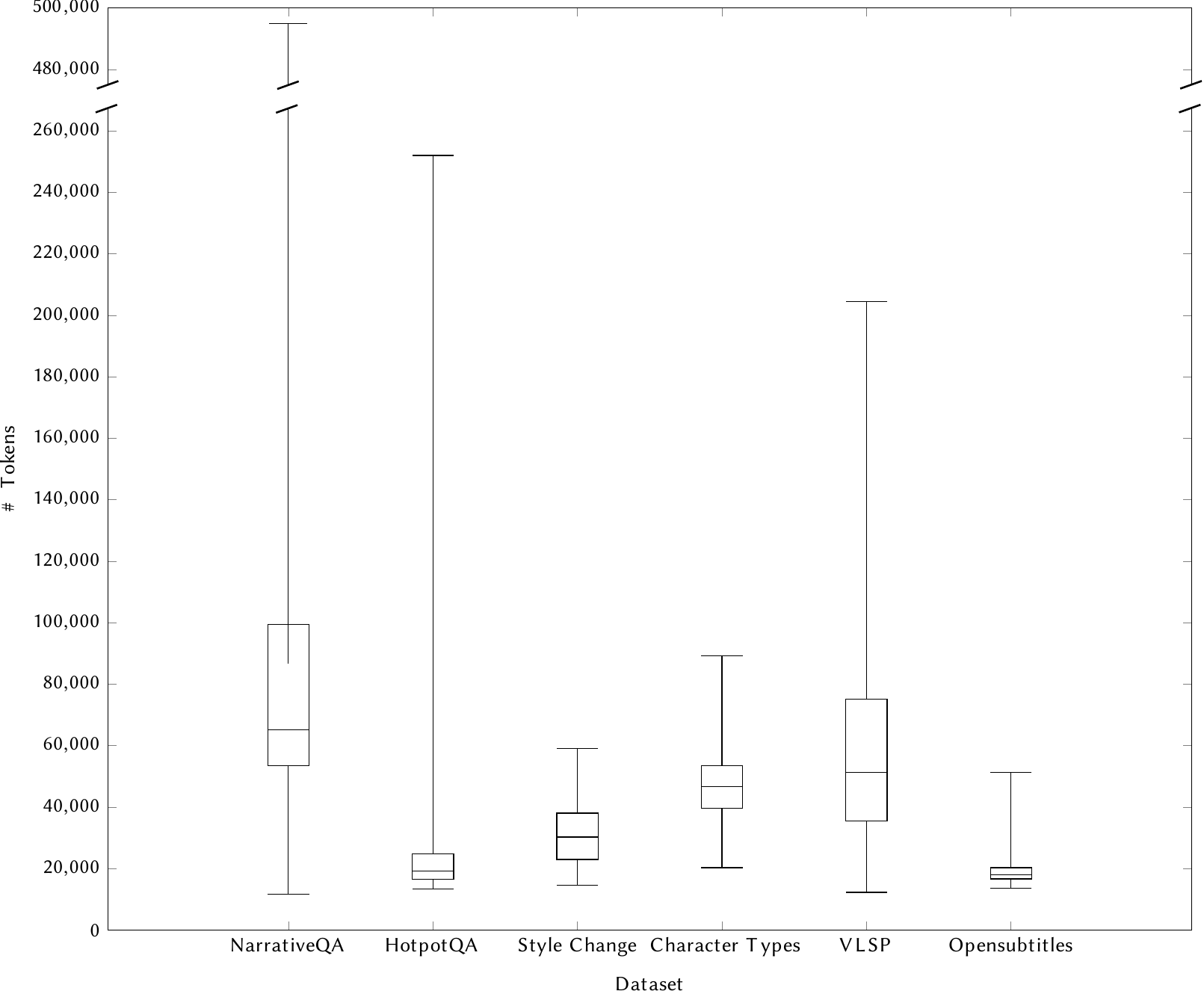}
    \caption{Dataset lengths (\#tokens)}
    \label{fig:ds_lengths}
\end{figure*}

\paragraph{NarrativeQA}
The NarrativeQA Reading Comprehension Challenge Dataset \cite{Kocisky_2018} consists of user-submitted questions regarding the plot of movies or novels. Annotators were only given access to a human-written plot summary to encourage general questions which require a full understanding of the narrative. When these are given to a question-answering system along with the full text of the narrative (either a movie script or the novel text), this is a test of reading comprehension. As each sentence the annotator read can summarise the plot of an entire chapter/scene of a book or movie, models evaluated on the full text must model the dependencies between multiple sections of the narrative. The majority of these documents are longer than our 10,00 token minimum - we simply filter out any documents shorter than this.

\paragraph{HotpotQA} The HotpotQA dataset consists of questions from crowd workers which require information from multiple Wikipedia articles in order to answer, thus testing the ability for models to perform multi-hop question answering. The data is commonly presented as a list of paragraphs containing relevant information plus a setting where the addition of 'distractor paragraphs' fully test the ability of the model to comprehend which information is relevant to the question asked. To transform this into a long document, we expand each paragraph with its full Wikipedia page as well as adding additional distractor articles from similar topics (randomly chosen from links on the existing pages) in order to meet the 10,000 token minimum length requirement for this benchmark. These articles are shuffled and concatenated to form the model input.

\paragraph{Character Archetype Classification} We introduce a character archetype classification dataset based on the methodology of \newcite{Skowron_2016}. For this dataset, each example consists of a movie script along with a named character and the task is to classify whether the character is a Hero/Protagonist or Villain/Antagonist based on understanding their role in the narrative.

To gather this data we first pick scripts from the web following \newcite{Kocisky_2018}\footnote{Primarily from the Internet movie script database: \url{www.imsdb.com}}, these are then matched with summaries of the plot from Wikipedia using a combination of matching names and titles with additional manual verification.

We extract character name candidates using a number of methods. Firstly, many of the scripts use the common format where the character name is given before each line of dialogue (e.g. `HARRY: Where have they gone?'). Secondly, the Wikipedia pages for many films include a list of characters that can be parsed and matched with the script. We then filter these character name candidates to only include those that also appear in the plot summary, eliminating false matches as well as some minor characters who don't impact the plot.

Annotators on Amazon Turk were given a description of the task and the character types of \newcite{Skowron_2016}. They were then provided with the plot summary and asked to select the character type for each character name candidate extracted previously. Multiple annotators were used on each example to ensure the accuracy of the labels. From this process we eliminated the character types Mentor, Sidekick, and Spouse as there was a lot of disagreement between these classes, perhaps due to our use of movies of all genres rather than the limited set of action movies used by \newcite{Skowron_2016}. 

\paragraph{Open Subtitles}
The Open Subtitles corpus \cite{Lison_2018OpenSubtitles}\footnote{\url{opus.nlpl.eu/OpenSubtitles2016.php}} consists of aligned subtitles from movies and TV shows from the website \url{opensubtitles.org} in 60 languages and can be used for machine translation. Importantly rather than individual lines, the data consists of the subtitles for an entire individual movie or tv show, many of these being very long files and we filter to remove any document with less than 10,000 tokens. 

One of the most common mistakes made by models which don't consider document-level context is the mistranslation of pronouns. For example, in the English phrase ``It is cold.'', the pronoun ``it'' should be translated differently depending on if the proceeding context was ``The ice formed at night'' or ``The camera was left outside''. For this reason we make use of the English-German pairs used by the ContraPro annotation of open subtitles \cite{muller2018large} which explicitly tests a model's ability to disambiguate these possibilities. This will allow future use of this contrastive evaluation, where models are asked to pick between the two possible translations and must show knowledge of the context.

\paragraph{AO3 style change detection}
Style change detection is the task of identifying the points where the author changes in a document constructed from the work of multiple authors. The PAN 2021 Style Change detection shared task \cite{Zangerle_2021Overview} introduced this task, forming documents from StackExchange comments to produce a challenging dataset with multiple increasingly difficult subtasks. Task 1 is a binary classification task to identify whether the document contains multiple authors or just a single author. Task 2 is to identify the points where the style changes, and Task 3 is to assign each paragraph uniquely to an author out of the number of authors in the document. As Task 3 is the most challenging (and answers the other two can be constructed from the output of task 3), we only report scores for it in our benchmark.

To extend this approach for a long document dataset, we instead use stories contributed to the fanfiction website \textit{Archive of Our Own}, which contains a large number of different works submitted by fans of popular films, tv, game, and book characters. 

We use stories written about four of the most popular character relationships: Sherlock Holmes \& John Watson, Castiel \& Dean Winchester, Steve Rogers \& Tony Stark, and Draco Malfoy \& Harry Potter. Since we want this task to be a test of style change detection and not topic change detection, each constructed document only contains paragraphs taken from the same relationship to ensure that the difference between sections is the author style not the topic. Additionally, we reserve the ``Draco Malfoy \& Harry Potter'' documents for the test set.
After downloading, we clean the data by removing any images and special formatting characters, then split the stories into paragraphs, removing any with less than 100 characters.

To construct the style change detection documents, we first randomly choose first randomly choosing a minimum document length (10,000-30,000) and a number of authors (1-4) with 50\% of documents having a single author. We then assign authors to the document, randomly partition the lengths for each section, and finally draw, shuffle, and concatenate paragraphs to form the complete text.

\paragraph{Very Long Scientific Papers (VLSP)}
We follow the process of the Scientific papers \cite{Cohan_2018} summarization dataset, extracting papers from the open-access preprint server Arxiv.org using the both the arxiv short abstract and the one included in the thesis (where available) as the reference summaries. In contrast to \newcite{Cohan_2018}, rather than removing very long documents, we explicitly include them - removing any document with less than 10,000 tokens. With this new filtering, the dataset mostly consists of theses which like the scientific papers dataset, have a regular structure consisting of multiple chapters. Due to the long time required to compile these large documents into text files, we only provide a small test set of 482 documents, so we train our models on the original smaller-length scientific papers dataset.

\paragraph{
}A box-plot of lengths for all six tasks is shown in \autoref{fig:ds_lengths}. We can see that NarrativeQA has the longest documents, but all tasks have a median length between 10,000 and 100,000 tokens. 

\section{Baselines}
\begin{table*}[]
    \centering
    \begin{tabular}{cccccc}
    \toprule
    Task     & Bleu-1 & Bleu-4 & RougeL & Meteor & F1\\
    \midrule
    \multicolumn{6}{c}{\textbf{T5}} \\
    \midrule
    NarrativeQA & 17.67 & 0.55 & 19.03 & 3.36 \\
    HotpotQA & 28.11 &	13.63 &	27.61 &	4.46 \\
    Style Change & & & & &  26.49\\
    Character Type & & & & & 54.01\\
    VLSP & 28.85 &	0.84 &	16.55 &	7.98 \\
    OpenSubtitles & 34.07 &	1.63 &	35.35 &	38.53 \\
    \midrule\midrule\multicolumn{6}{c}{\textbf{Longformer}} \\ \midrule
    NarrativeQA & 19.84 &	0.62 &	22.09 &	4.52 \\
    HotpotQA & 30.38 & 16.76 & 30.49 & 4.98 \\
    Style Change & & & & & 28.17\\
    Character Type  & & & & & 82.58 \\
    VLSP & 46.74 &	3.05 &	19.52 &	9.58 \\
    OpenSubtitles & 22.74 &	0.20 &	22.17 &	22.95\\

    \bottomrule
    \end{tabular}
    \caption{T5 and Longformer results on the benchmark}
    \label{tab:main_results}
\end{table*}


In this section we describe the models we evaluate on the MuLD benchmark. We experiment firstly with a model based on a 'standard' transformer architecture: T5, an encoder-decoder network which was pretrained on multiple text-to-text tasks and can take a maximum input length of 512 tokens \cite{Raffel_2020}. We compare this to the Longformer model: an 'efficient transformer' which uses sliding window attention (with global attention on some predetermined tokens) to supports up to 4096 tokens (or 16,384 in the encoder-decoder variant) allowing us to see the benefits of using longer contexts for the benchmark \cite{Beltagy_2020}.

As it is not possible to directly feed the entirety of many long documents into models on reasonable hardware (even with longformer), we use chunking techniques to divide, solve, and recombine parts of the input. We use the following methods:
\begin{itemize}
    \item\textbf{NarrativeQA/HotpotQA} - We follow the approach of \newcite{Kocisky_2018} by first dividing the document into chunks of 200 tokens. We then calculate the cosine similarity between the TF-IDF representations of each chunk and the question text in order to score every chunk. Using this metric, we pick the top 10 chunks and concatenate them together to pass into the model along with the question input.
    \item\textbf{Open Subtitles} - For a simple baseline, we split the document into regular chunks while respecting line breaks (we never end a chunk in the middle of a line). Each chunk is then passed into the model to be translated, and all the translated chunks are concatenated together to form the translated document.
    \item\textbf{Style Change Detection} - We use the methodology of \newcite{Zhang_2021style}, by training a classifier on paragraph pairs with the target of predicting whether the two paragraphs are by the same author. For the subtask we report (task 3), we use the methodology of \newcite{Strom_2021multi}: we assign the first paragraph to the first author, then check if the next paragraph is similar to the previous one. If so we assign this to the author of the most similar paragraph if over some threshold. If not we add a new author. Although we only report the score for task 3, this classifier could also be used to solve all the remaining two subtasks: for task 1, pass consecutive paragraph pairs into the model and output "single-author" if all paragraphs pairs are by the same author, otherwise classify the document as "multi-author". For task 2 simply append all the model outputs together to form a list of style changes. 
    \item\textbf{Character Archetype detection} - We select chunks containing the first mention, last mention, and most frequent mention of the character in concern. These are concatenated, passed to the model, and used to predict the character type class.
    \item\textbf{VLSP} - We split the document into chunks based on the article section headings and summarize the text from the first introduction section of the thesis onwards (ignoring contents, list of tables, list of figures sections).
\end{itemize}

Following \cite{Kocisky_2018}, we evaluate tasks which require text-generation (QA, Summarization, Translation) using Bleu-1, Bleu-4 \cite{Papineni_2002}, Meteor \cite{Denkowski_2011}, and Rouge-L \cite{Lin_2004}, using multiple references where these are available. For the style change detection and character type classification tasks, we simply report the F1-score.
\section{Results and Discussion}

The results for each of the benchmark tasks are presented in \autoref{tab:main_results} for both the T5 and Longformer models.

The Longformer model consistently outperforms the T5 model across many of the tasks, suggesting that models which are able to make use of a longer context perform well on our benchmark. 

Both models find the NarrativeQA dataset more challenging than HotpotQA, which we hypothesise is due to its longer average length, and the higher complexity of narrative understanding involved in NarrativeQA in contrast to the factual Wikipedia data of HotpotQA which typically involves only a limited number of hops. Additionally, each HotpotQA question commonly involves understanding only a small number of sentences (even though these may be widely distributed throughout the document).

The T5 model also outperforms Longformer on the OpenSubtitles translation task. We suggest that this is due to the challenge of having to output a much longer sequence (512 vs 4096) tokens is greater than the benefit gained on the few lines where the correct translation depends on context more than 512 tokens away (\newcite{muller2018large} find that in around half of cases, the antecedent is in the previous sentence, and very few are more than 3 sentences away).


\section{Conclusion}
To enable the evaluation of long document models, we introduce MuLD: a benchmark of varied NLP tasks where each document consists of more than 10,000 tokens. The six tasks in our benchmark are created by filtering, extending, or modifying existing NLP tasks and are designed to require a long context for high performance.

We evaluate simple chunking-based baselines, and find that the Longformer model is able to outperform the T5 model suggesting our benchmark is a good test for the ability of models to make use of longer contexts. 

We believe that the technique explored in this work of augmenting and extending existing `short document' datasets, can be applied to many other NLP tasks. As the performance of efficient transformers improves, we anticipate the need to update this benchmark with more challenging tasks. While we are focused on creating a benchmark which tests a model's ability to solve real-world long document tasks, we also expect improvements in the efficiencies of the models themselves which may make datasets with more than 100,000 tokens necessary which may require a fundamentally different approach to creating long document datasets. We leave for future work both the development of improved chunking methods, and more efficient transformers which make such methods unnecessary. 

We hope that the MuLD benchmark will encourage this further research into efficient models for long document NLP. To this end we provide the data, baseline models, and other code at \url{www.github.com/ghomasHudson/muld}.
\section{Bibliographical References}\label{reference}

\bibliographystyle{lrec2022-bib}
\bibliography{lib}

\label{lr:ref}

\onecolumn
\section*{Appendix}
We present examples from the MuLD benchmark to give the reader a sense of the tasks included (We have added ellipsis for brevity):

\subsection*{NarrativeQA}
\textbf{Input:}
\begin{lstlisting}
How is Oscar related to Dana?
[...]EXT.  MANHATTAN ISLAND - DAY
A high AERIAL SHOT of the island features the Statue of Liberty prominently in the
foreground then TRAVELS ACROSS the harbor, OVER the Battery and Lower Manhattan to 
Greenwich Village[...]
						DANA
				 (exasperated)
			Frank, do you think you could give me a hand
			with these bags?

						FRANK
			I'm not a doorman, Miss Barrett.  I'm a
			building superintendent[...]
\end{lstlisting}
\textbf{Output:}
\begin{lstlisting}
her son
\end{lstlisting}
\subsection*{HotpotQA}
\textbf{Input:}
\begin{lstlisting}
Were Scott Derrickson and Ed Wood of the same nationality? Doctor Strange is a
2016 American superhero film based on the Marvel Comics character of the same name. 
Produced by Marvel Studios and distributed by Walt Disney Studios Motion Pictures, 
it is the 14th film in the Marvel Cinematic Universe (MCU)[...]
Scott Derrickson (born July 16, 1966) is an American filmmaker. He is best known for 
directing the films The Exorcism of Emily Rose, Sinister, Deliver Us from Evil, and 
Doctor Strange[...]
Edward Davis Wood Jr. (October 10, 1924 - December 10, 1978) was an American 
filmmaker, actor, and author[...]
\end{lstlisting}
\textbf{Output:}
\begin{lstlisting}
Yes
\end{lstlisting}
\subsection*{Style Change}
\textbf{Input:}
\begin{lstlisting}
John's stomach had a new home again. It now settled into his throat, painfully dry for want of the man before him.
"Which story would you like to hear, my dear?" John's fingers looped in and around the curls so delicately, a lesser effort wouldn't have moved the hair at all. He thought of the beautiful mind beneath these curls. Once storing every memory like a computer and now weakened with age- but always beautiful.
John gulped down a lump of panic that crawled up his throat. What did he know? "Right, but I still reckon I'll keep pretending you don't."
"John, when we need to pretend to be a couple, things like spontaneous physical contact are expected to maintain-"
"Don't mention it," Sherlock responded with a soft passion he couldn't keep out of his voice. The words came out in a murmur.
-" he continued, delighted when John joined in to harmonize flawlessly in the latter half of phrase[...]
\end{lstlisting}
\textbf{Output:}
\begin{lstlisting}
1,1,1,1,1,1,1,[...],2,2,1,1,[...],3
\end{lstlisting}
\subsection*{Character Types}
\textbf{Input:}
\begin{lstlisting}
Indiana Jones
[...]
                INDY
        No.

Barranca looks evilly at Indy's hand upon him.
Indy releases him and smiles in a friendly way.

                INDY
        We don't need them.

Satipo watches this confrontation with 
some concern.

                BARRANCA
        I do not carry supplies.

                INDY
        We'll leave them. Once we've got it, 
        we'll be able to reach the plane by 
        dusk[...]
\end{lstlisting}
\textbf{Output:}
\begin{lstlisting}
Hero
\end{lstlisting}
\subsection*{VLSP}
\textbf{Input:}
\begin{lstlisting}
## Chapter 1 Basic Definitions and Concepts

### 1.1 Basics of simplicial complexes

Let @xmath , and @xmath denote the subsets of @xmath of size @xmath . A
collection @xmath of subsets of @xmath is called a (finite abstract)
simplicial complex if it is closed under inclusion, i.e. @xmath implies
@xmath . Note that if @xmath is not empty (which we will assume from now
on) then @xmath . The @xmath -th skeleton of @xmath is @xmath . The
elements of @xmath are called faces ; those in @xmath have dimension i .
The @xmath -dimensional faces are called vertices , the @xmath
-dimensional faces are called edges and the maximal faces with respect
to inclusion are called facets . If all the facets have the same
dimension, @xmath is pure . The @xmath - vector (face vector) of @xmath
is @xmath where @xmath . The dimension of K is @xmath ; e.g. a
1-dimensional simplicial complex is a simple graph. The @xmath -
polynomial of @xmath is @xmath[...]
\end{lstlisting}
\textbf{Output:}
\begin{lstlisting}
This thesis focuses on algebraic shifting and its applications to f-vector
theory of simplicial complexes and more general graded posets. In particular,
several approaches and partial results concerning the g-conjecture for
simplicial spheres are presented here.
\end{lstlisting}

\subsection*{OpenSubtitles}
\textbf{Input:}
\begin{lstlisting}
1957 was a big year.
The Russians put that Sputnik into outer space.
The Dodgers played their last game at Ebbets Field and said goodbye to Brooklyn.
That guy, he shot Frank Costello in the head and missed.
The Gallo brothers whacked Albert Anastasia in that barbershop.
It was total chaos[...]
\end{lstlisting}
\textbf{Output:}
\begin{lstlisting}
1957 war ein bedeutendes Jahr.
Die Russen schossen ihren Sputnik ins All.
Die Dodgers spielten zum letzten Mal in Ebbets Field und sagten Brooklyn Adieu.
Dieser Kerl schoss auf Frank Costello und verfehlte ihn.
Die Gallo-Bruder legten Albert Anastasia beim Friseur um.
Es war totales Chaos[...]
\end{lstlisting}
\end{document}